\documentclass[11pt]{article}
\usepackage[utf8]{inputenc}
\usepackage{times}
\usepackage{microtype}
\usepackage{amsmath,amssymb}
\usepackage{graphicx}
\usepackage{booktabs}
\usepackage{algorithm}
\usepackage{algpseudocode}
\usepackage{hyperref}
\usepackage{enumitem}
\usepackage{caption}
\usepackage{subcaption}
\usepackage{multirow}
\usepackage{geometry}
\usepackage{xcolor}
\usepackage{natbib}
\title{LLMRank: Understanding LLM Strengths for Model Routing}
\author{
    Shubham Agrawal \\
    Zeno AI \\
    \texttt{shubham@zenoai.tech}
    \and
    Prasang Gupta \\
    Independent Researcher \\
    \texttt{prasang.tix@gmail.com}
}

\date{September 2025}

\begin{document}
\maketitle

\begin{abstract}
The rapid growth of large language models (LLMs) with diverse capabilities, latency and computational costs presents a critical deployment challenge: selecting the most suitable model for each prompt to optimize the trade-off between performance and efficiency. We introduce \textbf{LLMRank}, a prompt-aware routing framework that leverages rich, human-readable features extracted from prompts, including task type, reasoning patterns, complexity indicators, syntactic cues, and signals from a lightweight proxy solver. Unlike prior one-shot routers that rely solely on latent embeddings, LLMRank predicts per-model utility using a neural ranking model trained on RouterBench, comprising 36,497 prompts spanning 11 benchmarks and 11 state-of-the-art LLMs, from small efficient models to large frontier systems. Our approach achieves up to 89.2\% of oracle utility, while providing interpretable feature attributions that explain routing decisions. Extensive studies demonstrate the importance of multifaceted feature extraction and the hybrid ranking objective, highlighting the potential of feature-driven routing for efficient and transparent LLM deployment.
\end{abstract}

\section{Introduction}

Large Language Models (LLMs) have rapidly advanced natural language processing, enabling strong performance across diverse tasks such as reasoning, summarization, and code generation. However, the proliferation of models with widely varying capabilities, costs, and latencies has introduced a fundamental deployment challenge: how to select the right model for a given query and parameters. Using a frontier model like GPT-5 for all inputs is prohibitively expensive, while relying exclusively on lightweight models such as Mistral-7B can sacrifice accuracy on complex queries. 

This motivates the task of \emph{LLM routing}, which dynamically assigns each prompt to the most suitable model in a pool to balance cost and performance. \citet{routerbench} showed that an oracle selector (always choosing the best model based on output) can outperform the best-performing model (GPT-4 at the time of their work) while reducing inference costs dramatically. A well-designed router can approach oracle-level performance. However, designing such a router remains a challenge. Current methods often treat prompts as opaque embeddings, limiting interpretability and generalization. Others optimize for either performance or cost in isolation, overlooking the need for flexible trade-offs. Furthermore, few approaches provide insight into \emph{why} a given routing decision is made. Finally, ever-evolving model pools require a smarter router that can seamlessly accommodate new models or remove outdated ones without retraining the entire system.

We introduce \textbf{LLMRank}, a feature-driven routing framework that addresses these limitations. LLMRank makes three key contributions:
\begin{itemize}[itemsep=2pt,topsep=3pt]
    \item \textbf{Human-interpretable feature extraction.} We develop a pipeline that derives explicit features from prompts, including task type indicators, linguistic and semantic complexity, domain signals, and proxy model predictions.
    \item \textbf{Hybrid ranking objective.} We train a neural router that combines pointwise utility prediction with pairwise ranking losses, enabling nuanced discrimination of model strengths across prompts.
    \item \textbf{Cost-aware routing.} We incorporate model cost directly into training, yielding flexible deployment strategies that trade accuracy for efficiency in a principled way.
\end{itemize}

Through experiments on RouterBench and related benchmarks, LLMRank achieves state-of-the-art routing quality at a fraction of the computational cost. Our feature attribution analysis further reveals intuitive patterns in model selection (e.g., complex reasoning vs. factual lookup), offering transparency often missing in prior routing approaches. 

\section{Related Work}

 The task of routing has attracted growing attention with the availability of many open and closed source LLMs. \citet{routerbench} introduced a dataset and benchmark standardizing evaluation of routing strategies across models and tasks. Early methods such as FrugalGPT \citep{frugalgpt} explored cascading strategies to cut costs while maintaining quality. More recent approaches propose increasingly sophisticated routers: Zooter \citep{zooter} distills reward models into routing policies, RadialRouter \citep{radialrouter} introduces structured query–model representations and achieves strong performance on RouterBench, Routoo \citep{routoo} predicts model performance with cost-aware selection, and TagRouter \citep{tagrouter} leverages tag-based representations for open-domain text generation tasks. Training-free strategies such as Eagle \citep{eagle} approximate routing via Elo-style scoring, while CARGO \citep{cargo} leverages uncertainty-aware regression for category-specific routing. Router-R1 \citep{routerr1} combines routing with reinforcement learning and multi-round aggregation.  

A key design choice is how to represent the input prompt. While most work relies on dense embeddings, some explore richer structures: RadialRouter uses radial-former interactions between query and model; IRT-Router \citep{irtrouter} applies psychometric Item Response Theory to capture query difficulty and model ability, providing interpretability and cold-start robustness. TagRouter \citep{tagrouter} complements this trend by using discrete tag representations for routing. Our work differs by systematically extracting explicit, human-readable features, demonstrating that they can rival or complement embedding-based approaches.

Beyond routing, ensembles of multiple LLMs have been studied extensively. Surveys such as \citet{llmensemble_survey} and \citet{ensemble_survey} categorize ensemble strategies into before-inference (routing, selection), during-inference (mixtures, token-level fusion), and after-inference (output aggregation). These surveys highlight routing as a critical subproblem with open challenges in interpretability, generalization, and cost-awareness—precisely the gaps LLMRank aims to address.

Feature engineering played a central role in pre-deep-learning NLP \citep{feature_engineering}. More recently, LLM-based feature generation has been shown effective for interpretable machine learning \citep{llmfeatures}. Our work adapts this paradigm to LLM routing: extracting meaningful features from prompts and leveraging them in a hybrid neural ranking model.

\section{Problem Formulation}

Following RouterBench \cite{routerbench}, we formalize the routing problem as follows. Let $\mathcal{L} = \{\text{LLM}_1, \ldots, \text{LLM}_m\}$ denote a set of $m$ candidate models, and let $\mathcal{D} = \{(x_i, y_i)\}_{i=1}^N$ represent a dataset of prompts and their ground-truth outputs. For each prompt $x$ and model $j \in [m]$, we have:
\begin{itemize}[itemsep=1pt,topsep=2pt]
    \item Quality score $q_j(x) \in [0, 1]$: task-specific performance (e.g., accuracy for classification, BLEU for generation)
    \item Cost $c_j$: dollar cost per inference (including API fees and amortized compute as provided in the dataset)
\end{itemize}

A router $R: \mathcal{X} \to [m]$ maps each prompt to a model index. The router's expected quality and cost over the dataset are:
\begin{align}
    Q_R &= \mathbb{E}_{x \sim \mathcal{D}}[q_{R(x)}(x)] \\
    C_R &= \mathbb{E}_{x \sim \mathcal{D}}[c_{R(x)}]
\end{align}

    The goal is to learn a router that maximizes quality while respecting cost constraints, or equivalently, maximizes the cost-adjusted utility:
\begin{equation}
    U_R = Q_R - \lambda \cdot C_R
\end{equation}
where $\lambda \geq 0$ is a hyperparameter controlling the cost-quality trade-off.

Following the methodology of \citet{radialrouter}, we evaluate different values of the trade-off parameter $\lambda$ by comparing three scenarios: LLMRank-Perf ($\lambda=0$), LLMRank-Balanced ($\lambda=10^3$), and LLMRank-Cost ($\lambda=10^5$). In our setup, for utility, we report costs on a per-query basis, rather than normalizing to cost per 1K tokens as done in prior work.
This setup aligns with the RouterBench framework, where $\lambda$ plays a role analogous to the willingness-to-pay (WTP) parameter. 

\section{The RouterBench Dataset}

RouterBench~\cite{routerbench} is a large-scale evaluation framework for model routing that aggregates prompts from diverse benchmarks and provides precomputed outputs from multiple LLMs. We adopt the zero-shot variant (`routerbench\_0shot.pkl`), which contains 36,497 prompts with associated responses, performance metrics, and cost estimates. Each entry includes:
\begin{itemize}[itemsep=1pt,topsep=2pt]
    \item Natural language prompts spanning 11+ benchmark categories,
    \item Candidate responses from 11 state-of-the-art LLMs,
    \item Quality scores (binary or continuous, depending on task),
    \item Per-inference costs based on API pricing,
    \item Oracle labels denoting the cost-quality optimal model choice (cheapest model with the highest performance for that task).
\end{itemize}

\subsection{Dataset Composition}
After filtering low-resource categories ($<$50 samples), prompts in Chinese language, and discarding prompts unsolved by all models, the dataset contains 34,623 prompts in 10 categories. Table~\ref{tab:dataset_stats} summarizes the distribution, dominated by reasoning-heavy benchmarks such as MMLU, HellaSwag, and GSM8K.

\begin{table}[ht]
\centering
\caption{RouterBench composition after preprocessing.}
\label{tab:dataset_stats}
\begin{tabular}{lrr}
\toprule
\textbf{Benchmark} & \textbf{Samples} & \textbf{Percentage} \\
\midrule
MMLU (57 subjects) & 13,408 & 38.7\% \\
HellaSwag & 9,800 & 28.3\% \\
GSM8K (Grade School Math) & 7,450 & 21.5\% \\
ARC-Challenge & 1,456 & 4.2\% \\
WinoGrande & 1,267 & 3.7\% \\
MBPP (Code) & 370 & 1.1\% \\
Consensus & 362 & 1.0\% \\
Abstract2Title & 254 & 0.7\% \\
BiasDetection & 176 & 0.5\% \\
MT-Bench & 80 & 0.2\% \\
\midrule
\textbf{Total} & \textbf{34,623} & \textbf{100\%} \\
\bottomrule
\end{tabular}
\end{table}

\subsection{Model Pool and Oracle Distribution}
RouterBench includes responses from 11 LLMs spanning diverse sizes and cost profiles (Table~\ref{tab:model_dist}). Oracle selections are dominated by small-to-medium models, with 52.8\% of prompts optimally handled by models under 20B parameters. Larger models are chosen only for a minority of difficult cases, underscoring their complementary role. This distribution highlights the value of adaptive routing: it enables substantial cost reductions by defaulting to efficient models while reserving expensive capacity for queries that truly benefit from it. Further details and analysis of the dataset can be found in Appendix~\ref{appendix:additional_analysis}.

\begin{table}[ht]
\centering
\caption{Model pool with oracle frequencies and total cost incurred.}
\label{tab:model_dist}
\begin{tabular}{lrr}
\toprule
\textbf{Model} & \textbf{Oracle Freq.} & \textbf{Total Cost(\$)}\ \\
\midrule
mistralai/mistral-7b-chat & 28.9\% & 0.44 \\
WizardLM/WizardLM-13B-V1.2 & 23.9\% & 0.55 \\
mistralai/mixtral-8x7b-chat & 17.3\% & 0.77 \\
zero-one-ai/Yi-34B-Chat & 15.2\% & 1.05 \\
claude-instant-v1 & 4.5\% & 0.39 \\
gpt-3.5-turbo-1106 & 2.8\% & 0.31 \\
gpt-4-1106-preview & 2.0\%  & 2.40 \\
meta/llama-2-70b-chat & 1.9\% & 0.20 \\
meta/code-llama-34b-chat & 1.4\% & 0.14 \\
claude-v1 & 1.1\%  & 1.13\\
claude-v2 & 0.9\% & 1.09\\
\bottomrule
\end{tabular}
\end{table}

\section{LLMRank Method}

LLMRank is designed as a cost-aware routing framework that learns to predict model utilities and select the most appropriate model for a given prompt. The method integrates three complementary components: feature extraction (Section \ref{sec:feature-extraction}), a neural ranking model that maps features to per-model utility scores (Section~\ref{sec:ranking-model}), and cost-aware inference-time routing that balances performance with efficiency (Section \ref{sec:cost-aware-inference}). Together, these components enable LLMRank to generalize routing decisions across diverse tasks while respecting user-defined cost-quality trade-offs.

\subsection{Feature Extraction}
\label{sec:feature-extraction}

Each prompt is represented as a fixed-dimensional feature vector $\mathbf{f}(x) \in \mathbb{R}^{N}$, where $N$ are the total features generated from the Feature extraction module. The features combine multiple complementary aspects of the task. Difficulty is captured through categorical and continuous scores that indicate overall hardness and structural complexity. The task type indicators encode whether the prompt requires common sense reasoning, scenario completion, multiple choice, or narrative understanding. Knowledge requirements reflect the need for world knowledge, temporal reasoning, context understanding, and domain specificity. The output format features specify expected answer types, such as single-character responses, free-form completions, or deterministic output. Scenario complexity is quantified through ambiguity and contextual difficulty measures, together with lexical cues such as 'what happens next' or probabilistic phrasing. Routing hints describe the model capabilities needed, including reasoning, context, or world knowledge. Finally, quality indicators capture prompt length (normalized) and language type.

\subsection{Neural Ranking Model}
\label{sec:ranking-model}

\begin{figure}[t]
    \centering
    \includegraphics[width=0.8\linewidth]{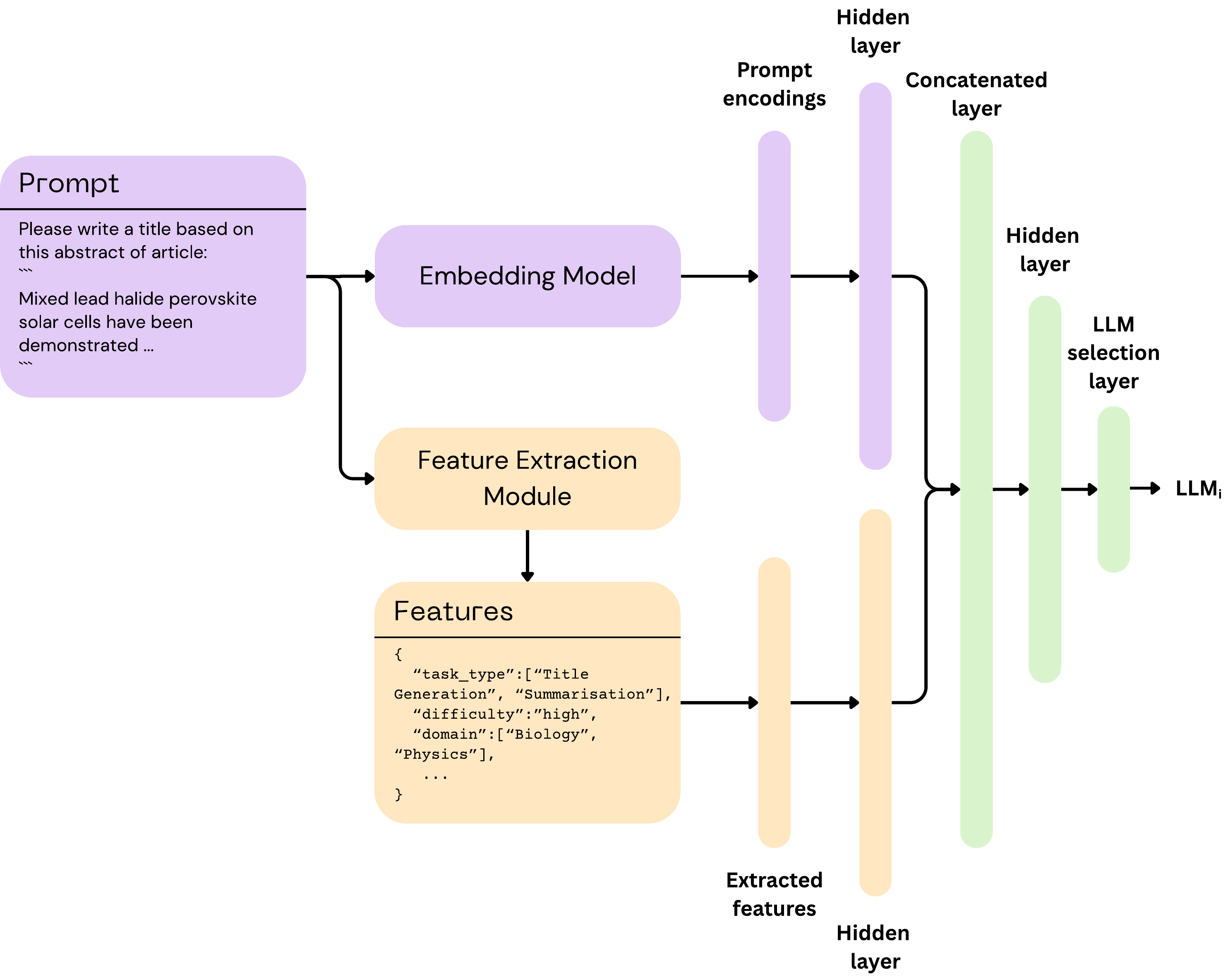}
    \caption{Architecture of the LLMRank ranking model. The feature extraction module encodes prompt-level signals, which are processed by a neural ranking network to produce cost-adjusted utility scores for candidate models.}
    \label{fig:model_arch}
\end{figure}

The ranking network $g_\phi: \mathbb{R}^d \to \mathbb{R}^m$ predicts utility scores for $m$ candidate models. The architecture diagram is shown in Figure~\ref{fig:model_arch} which represents the flow from an input prompt to its final corresponding set of utility scores.

\begin{align}
\text{Extracted features:}\quad
\mathbf{h}^{(j)}_1 &= \mathrm{ReLU}\!\left(\mathbf{W}^{(j)}_1 \mathbf{x}_j + \mathbf{b}^{(j)}_1\right)\\
\mathbf{z}_j &= \mathbf{W}^{(j)}_2\,\mathrm{Drop}_{0.1}\!\left(\mathbf{h}^{(j)}_1\right) + \mathbf{b}^{(j)}_2\\[4pt]
\text{Text encoder:}\quad
\mathbf{h}^{(t)}_1 &= \mathrm{ReLU}\!\left(\mathbf{W}^{(t)}_1 \mathbf{x}_t + \mathbf{b}^{(t)}_1\right)\\
\mathbf{z}_t &= \mathbf{W}^{(t)}_2\,\mathrm{Drop}_{0.1}\!\left(\mathbf{h}^{(t)}_1\right) + \mathbf{b}^{(t)}_2\\[4pt]
\text{Fusion:}\quad
\mathbf{c} &= [\mathbf{z}_j;\mathbf{z}_t]\\
\mathbf{h}^{(f)}_1 &= \mathrm{ReLU}\!\left(\mathbf{W}^{(f)}_1 \mathbf{c} + \mathbf{b}^{(f)}_1\right)\\
\mathbf{s} &= \mathbf{W}^{(f)}_2\,\mathrm{Drop}_{0.1}\!\left(\mathbf{h}^{(f)}_1\right) + \mathbf{b}^{(f)}_2
\end{align}

\noindent
Here $\mathbf{x}_j \in \mathbb{R}^{d_j}$ and \(\mathbf{x}_t \in \mathbb{R}^{d_t}\) are the extracted features and text embeddings, respectively; 
\([\,\cdot\,;\,\cdot\,]\) denotes concatenation; 
\(\mathrm{Drop}_{0.1}(\cdot)\) is dropout with rate \(0.1\) (applied only during training). 
Parameter shapes: \(\mathbf{W}^{(j)}_1 \in \mathbb{R}^{h \times d_j}\), \(\mathbf{W}^{(j)}_2 \in \mathbb{R}^{h \times h}\), 
\(\mathbf{W}^{(t)}_1 \in \mathbb{R}^{h \times d_t}\), \(\mathbf{W}^{(t)}_2 \in \mathbb{R}^{h \times h}\),
\(\mathbf{W}^{(f)}_1 \in \mathbb{R}^{h \times 2h}\), \(\mathbf{W}^{(f)}_2 \in \mathbb{R}^{m \times h}\).
The output \(\mathbf{s} \in \mathbb{R}^m\) contains the predicted utilities for the \(m\) candidate models.

\subsection{Training Objectives}
The router is trained to predict cost-adjusted utilities for each candidate model. 
For an input prompt $x$, the late-fusion encoder $g_\phi$ (Figure~\ref{fig:model_arch}) produces predicted utilities $\mathbf{s}(x) = g_\phi(\mathbf{f}(x)) \in \mathbb{R}^m$,
where $m$ is the number of candidate models. 
The ground-truth cost-adjusted utilities are given by
\begin{equation}
u_j(x) = q_j(x) - \lambda c_j ,
\end{equation}
where $q_j(x)$ is the task performance of model $j$, $c_j$ is its cost, 
and $\lambda$ is a tunable parameter controlling the cost–quality trade-off. The training objective combines pointwise regression with listwise ranking.

\paragraph{Pointwise utility regression.}
A mean squared error (MSE) loss penalizes deviations between predicted and target utilities:
\begin{equation}
\mathcal{L}_{\text{MSE}}
= \frac{1}{N}\sum_{i=1}^N \frac{1}{m}\sum_{j=1}^m
\big(s_j^{(i)} - u_j^{(i)}\big)^2 .
\end{equation}

\paragraph{Listwise ranking via KL divergence.}
To encourage correct relative ordering among models, 
we align predicted distributions with target distributions using KL divergence. 
With temperature $\tau=0.5$, define
\begin{equation}
p^{(i)} = \mathrm{softmax}\!\left(u^{(i)}/\tau\right), 
\quad 
\hat p^{(i)} = \mathrm{softmax}\!\left(s^{(i)}/\tau\right).
\end{equation}
The listwise loss is
\begin{equation}
\mathcal{L}_{\text{list}}
= \frac{1}{N}\sum_{i=1}^N
\mathrm{KL}\!\left(p^{(i)} \,\middle\|\, \hat p^{(i)}\right).
\end{equation}

\paragraph{Final objective.}
The overall training loss is the unweighted sum of the two terms:
\begin{equation}
\mathcal{L} = \mathcal{L}_{\text{MSE}} + \mathcal{L}_{\text{list}}.
\end{equation}

\subsection{Cost-Aware Inference}
\label{sec:cost-aware-inference}
At inference time, the router selects the model with the highest predicted utility:
\begin{equation}
R(x) = \arg\max_{j \in [m]} \, s_j(x).
\end{equation}
Since cost adjustments are integrated into the training targets $u_j(x)$, 
no explicit cost term is needed during inference. 
This design allows routers to be instantiated with different $\lambda$ values, 
yielding performance-first, balanced, or cost-first routing strategies. Additional details on the experimental setup are provided in Appendix~\ref{appendix:experimental_setup}.

\section{Results}

We evaluate LLMRank on the RouterBench test set, which contains 5,193 prompts spanning 10 benchmarks and responses from 11 candidate models. We compare three LLMRank configurations with different cost–quality trade-offs against state-of-the-art routing methods and single-model baselines. Results are reported in terms of average correctness (quality), total inference cost, and derived metrics like efficiency, cost ratio, and quality gap relative to the oracle.

\begin{table*}[ht]
\centering
\caption{Overall performance on RouterBench test set. Best results in \textbf{bold}, second-best \underline{underlined}. Efficiency measures the percentage of oracle utility achieved.}
\label{tab:main_results}
\begin{tabular}{lcccccc}
\toprule
\textbf{Method} & \textbf{Quality} & \textbf{Cost (\$)} & \textbf{Efficiency} & \textbf{Cost Ratio} & \textbf{Quality Gap} \\
\midrule
Oracle & 0.945 & 1.271 & 100.0\% & 1.00× & -- \\
\midrule
\textbf{LLMRank-Perf} ($\lambda=0$) & \textbf{0.843} & 5.784 & 89.2\% & 4.55× & 10.2\% \\
\textbf{LLMRank-Balanced} ($\lambda=10^3$) & 0.794 & 1.413 & 84.0\% & 1.11× & 15.1\% \\
\textbf{LLMRank-Cost} ($\lambda=10^5$) & 0.743 & \underline{0.832} & 78.6\% & 0.65× & 20.2\% \\
\midrule
RadialRouter$^{*}$ & \underline{0.816} & 6.759 & 86.3\% & 5.32× & 12.9\% \\
RouterDC$^{*}$ & 0.815 & 6.768 & 86.2\% & 5.33× & 13.0\% \\
\midrule
GPT-4-1106 (Best Single) & 0.812 & 17.237 & 85.9\% & 13.56× & 13.3\% \\
Mistral-7B (Cheapest) & 0.571 & \textbf{0.708} & 60.4\% & 0.56× & 36.4\% \\
\bottomrule
\end{tabular}

\vspace{2mm}
\footnotesize
$^{*}$Values for RadialRouter and RouterDC are directly taken from the radial router paper due to lack of implementation.
\end{table*}

Table~\ref{tab:main_results} summarizes the overall comparison. LLMRank-Perf achieves the highest quality among router-based methods, surpassing even the strongest single model (GPT-4-1106), while retaining 89.2\% of oracle utility. The cost-oriented variant, LLMRank-Cost, reduces inference cost to just 65\% of the oracle baseline while still maintaining 78.6\% efficiency. LLMRank-Balanced provides the best trade-off: it achieves only 1.8\% lower quality than GPT-4-1106 but at nearly 12$\times$ lower cost. These results demonstrate that routing not only improves over single-model baselines but also delivers substantial efficiency gains. The use of open-source models further amplifies this advantage due to their favorable pricing relative to proprietary systems.

The three LLMRank variants trace out distinct points along the cost–quality Pareto frontier. LLMRank-Perf prioritizes accuracy and approaches oracle-level quality, LLMRank-Cost minimizes inference expenditure while retaining reasonable performance, and LLMRank-Balanced achieves a middle ground that is particularly competitive in practical settings. This flexibility demonstrates that routing can be tuned to deployment requirements, offering practitioners a principled way to trade accuracy for cost depending on application constraints. We next analyze how these trade-offs manifest across individual benchmarks.

\begin{table*}[ht]
\centering
\caption{Performance breakdown by benchmark. Values show average quality scores for each method.}
\label{tab:benchmark_performance}
\begin{tabular}{lcccccccccc}
\toprule
\textbf{Benchmark} & \textbf{Oracle} & \textbf{LLMRank} & \textbf{LLMRank} & \textbf{LLMRank} & \textbf{Radial}$^{*}$ & \textbf{GPT-4} &  \textbf{Mistral} \\
 & & \textbf{-Perf} & \textbf{-Balanced} & \textbf{-Cost} & \textbf{Router} & & \textbf{-7B} \\
\midrule
Abstract2Title & 1.000 & 1.000 & 0.999 & 0.996 & -- & 1.000 & 0.996 \\
ARC-Challenge & 1.000 & 0.976 & 0.956 & 0.865 & 0.956 & 0.971 & 0.391 \\
BiasDetection & 1.000 & 0.563 & 0.503 & 0.483 & -- & 0.608 & 0.097 \\
Consensus & 0.934 & 0.701 & 0.685 & 0.652 & -- & 0.570 & 0.642 \\
GSM8K & 0.749 & 0.672 & 0.578 & 0.516 & 0.667 & 0.659 & 0.412 \\
HellaSwag & 1.000 & 0.914 & 0.846 & 0.793 & 0.906 & 0.860 & 0.261 \\
MBPP & 1.000 & 0.744 & 0.622 & 0.523 & 0.695 & 0.792 & 0.397 \\
MMLU & 1.000 & 0.852 & 0.827 & 0.764 & 0.816 & 0.850 & 0.266 \\
MT-Bench & 0.919 & 0.806 & 0.795 & 0.738 & -- & 0.806 & 0.524 \\
WinoGrande & 1.000 & 0.894 & 0.848 & 0.791 & 0.855 & 0.819 & 0.524 \\
\bottomrule
\end{tabular}

\vspace{2mm}
\footnotesize 
$^{*}$Due to lack of implementation details and possible data modifications, 
results for Radial Router are reported directly from the original paper. 
Thus, the test dataset may differ, making this an approximate comparison.
\end{table*}

Table~\ref{tab:benchmark_performance} reports quality scores across individual benchmarks. LLMRank consistently outperforms baseline routers on large-scale tasks such as MMLU, HellaSwag, and GSM8K, reflecting its ability to exploit prompt semantics for task-specific routing. On smaller benchmarks (e.g., MT-Bench, Bias Detection, Consensus), performance margins narrow, as limited training signals constrain generalization and the router tends to fall back on the strongest single models. These results indicate that while LLMRank scales effectively across diverse tasks, benchmark size and quality remain key factors in routing reliability.

\section{Discussion}

A central question in model routing is: \emph{what makes a good router?} 
An effective system must balance multiple objectives. It should adapt to prompt-level characteristics rather than relying on aggregate statistics, scale gracefully as new models or datasets emerge, and remain robust under noisy or imperfect benchmark labels. In addition, interpretability is crucial for trust and debugging, while practical efficiency requires that routing overhead not outweigh the benefits of model selection. 

Most existing approaches fall short of these goals. In cost-unaware settings, routers often collapse to trivial solutions, sending nearly all queries to a single high-performing model (e.g., GPT-4), which defeats the purpose of routing and is unrealistic in deployment. Furthermore, many methods tightly couple the router to model identities, requiring full retraining when new models are introduced. Their black-box nature further limits transparency, leaving practitioners with little insight into why specific routing decisions are made.

LLMRank addresses these gaps through a semantic-aware, feature-driven design. By incorporating prompt semantics and multiple feature signals, it distinguishes between queries that require advanced reasoning and those that can be solved by smaller models, thereby maintaining diversity in routing even under performance-focused objectives. Its modular architecture further supports seamless integration of new models and datasets: adding a model requires only capability scoring and an appended weight vector, without architectural changes or full retraining. This design significantly reduces integration time while keeping the router up to date with evolving model pools and benchmarks. The multi-signal approach also enhances robustness to label noise, while feature attributions provide interpretable routing decisions, helping practitioners understand both routing behavior and model capabilities. Beyond text LLMs, the framework is extendable to multimodal generative models, offering a path toward general-purpose routing across AI~systems.

Despite these advantages, LLMRank has limitations. Its reliance on proxy models for feature extraction introduces bias and can affect stability, while the additional overhead of proxy inference may hinder adoption in latency-sensitive settings. Moreover, the current design does not explicitly capture dependencies across related or sequential prompts, limiting its effectiveness in multi-turn or session-level routing. These challenges, however, are more tractable than the systemic issues in prior approaches and open promising directions for future research. The authors are actively working on extensions that address
model, dataset, and framework limitations, which will be presented in subsequent work.

Our evaluation also revealed external limitations in current benchmarks and frameworks. Many datasets rely on large models as evaluators, which can introduce noise, and are dominated by synthetic or academic prompts rather than organic human queries. Variability across prompt phrasing, context, or multi-turn interactions is rarely captured, leading to potential overestimation of model reliability. Finally, most frameworks overlook deployment factors such as latency, throughput, or infrastructure cost, which are critical in practical settings. Addressing these dataset and framework gaps will be essential for fair and realistic evaluation of routing methods in the future.

\section{Conclusion}

We present LLMRank, a feature-driven approach to LLM routing that combines interpretable prompt analysis with learned routing policies. Through comprehensive experiments on RouterBench, we demonstrate that explicit feature engineering, when combined with modern neural ranking techniques, can achieve strong cost–quality trade-offs while providing interpretable routing decisions. LLMRank achieves over 89.2\% of oracle utility in balanced settings, outperforming existing baselines and open-source routing models.

Our analysis of existing routing methods reveals key limitations. Many approaches converge to trivial policies that simply learn to send most queries to the single best-performing model at a given cost, rather than adapt to prompt-level variation. In addition, they often require retraining the entire router when new models enter the ecosystem, and their architectures become increasingly opaque, offering little interpretability for benchmarking or evaluation. LLMRank addresses these issues through semantic-aware routing, modular design, and robust multisignalfeature integration.

Our work highlights the value of hybrid approaches that combine the interpretability of feature engineering with the flexibility of neural models. As the LLM ecosystem continues to expand, intelligent routing will become increasingly critical for practical deployments. LLMRank provides a foundation for building transparent, efficient and adaptable routing systems.

\bibliographystyle{plainnat}

\begin{thebibliography}{99}
\bibitem[Hu et~al.(2024)]{routerbench}
Hu, Qitian Jason and Bieker, Jacob and Li, Xiuyu and Jiang, Nan and Keigwin, Benjamin and Ranganath, Gaurav and Keutzer, Kurt and Upadhyay, Shriyash K.
\textit{RouterBench: A Benchmark for Multi-LLM Routing System}.
arXiv:2403.12031, 2024.

\bibitem[Jin et~al.(2024)]{radialrouter}
Jin, Ruihan and Shao, Pengpeng and Wen, Zhengqi and Wu, Jinyang and Feng, Mingkuan and Zhang, Shuai and Tao, Jianhua.
\textit{RadialRouter: Structured Representation for Efficient and Robust Large Language Models Routing}.
arXiv:2506.03880, 2024.

\bibitem[Chen et~al.(2023)]{frugalgpt}
Chen, Lingjiao and Zaharia, Matei and Zou, James.
\textit{FrugalGPT: How to Use Large Language Models While Reducing Cost and Improving Performance}.
arXiv:2305.05176, 2023.

\bibitem[Domingos(2012)]{feature_engineering}
Domingos, Pedro.
\textit{A Few Useful Things to Know About Machine Learning}.
Communications of the ACM, 2012.

\bibitem[Lu et~al.(2024)]{zooter}
Lu, Yufei and Gao, Tianyu and Zeng, Zhiyuan and Jiang, Zhengbao and Chen, Danqi.
\textit{Zooter: Reward-Guided Routing among Large Language Models}.
NAACL, 2024.

\bibitem[Zhang et~al.(2024)]{eagle}
Zhang, Yifan and Sun, Haoran and Jin, Bowen and Liu, Zhiwei and Han, Jiawei.
\textit{Eagle: Efficient Training-Free Router for Multi-LLM Inference}.
NeurIPS, 2024.

\bibitem[Xie et~al.(2025)]{cargo}
Xie, Junlin and Sun, Jianing and Qiao, Yu and Liang, Yuxuan.
\textit{CARGO: Category-Aware Routing with Gap-based Optimization}.
arXiv:2509.14899, 2025.

\bibitem[Wang et~al.(2025)]{routerr1}
Wang, Tianle and Wang, Xiaohui and Zhang, Hongyu and Li, Qian.
\textit{Router-R1: Reinforcement Learning for Multi-round Routing of Large Language Models}.
arXiv:2506.09033, 2025.

\bibitem[Li et~al.(2025)]{routoo}
Li, Zhiyuan and Bao, Keqin and Zhang, Jie and Liang, Yiming.
\textit{Routoo: Cost-Aware Routing via Performance Prediction for Multi-LLM Inference}.
OpenReview, 2025.

\bibitem[Liu et~al.(2025)]{irtrouter}
Liu, Zeyu and Liu, Qi and Sun, Xu.
\textit{IRT-Router: Psychometric Routing for Large Language Models}.
ACL, 2025.

\bibitem[Kliegr et~al.(2024)]{llmfeatures}
Kliegr, Tomáš and Sedivý, Jiří and Svátek, Vojtěch.
\textit{LLM-based Feature Generation from Text for Interpretable Machine Learning}.
Preprint, 2024.

\bibitem[Chen et~al.(2025a)]{llmensemble_survey}
Chen, Junchen and Zhou, Siyuan and Xu, Zhiqiang and Liu, Pengfei.
\textit{Harnessing Multiple Large Language Models: A Survey on LLM Ensemble}.
arXiv:2403.08384, 2025.

\bibitem[Zhang et~al.(2025)]{ensemble_survey}
Zhang, Wei and Wang, Shuo and Li, Yidong and Ma, Fenglong.
\textit{Ensemble Large Language Models: A Survey}.
Information, {\bf 16}(8): 688, 2025.

\bibitem[Chen et~al.(2025b)]{tagrouter}
Chen, Zhou and Wei, Zhiqiang and Bai, Yuqi and Xiong, Xue and Wu, Jianmin.
\textit{TAGROUTER: Learning Route to LLMs through Tags for Open-Domain Text Generation Tasks}.
arXiv:2506.12473, 2025.

\end{thebibliography}

\clearpage
\appendix

\section{Additional Analysis}
\label{appendix:additional_analysis}

\subsection{Oracle Routing Distribution}
Table~\ref{tab:routing_distribution} reports the distribution of oracle-selected models across benchmarks, highlighting distinct patterns in optimal model choice. For simpler tasks such as Abstract2Title, the oracle predominantly selects Mistral-7B, reflecting its low cost and sufficient accuracy. In contrast, other benchmarks exhibit a more diverse distribution, indicating that task complexity often necessitates leveraging a broader set of models.

\begin{table}[ht]
\centering
\caption{Routing distribution of oracle model selections across benchmarks. Values denote the percentage of prompts routed to each model.}
\label{tab:routing_distribution}
\resizebox{\textwidth}{!}{%
\begin{tabular}{lcccccccccccc}
\toprule
\textbf{Benchmark} & \textbf{WizardLM} & \textbf{Claude Inst.} & \textbf{Claude V1} & \textbf{Claude V2} & \textbf{GPT-3.5} & \textbf{GPT-4} & \textbf{CodeLlama} & \textbf{Llama 70B} & \textbf{Mistral} & \textbf{Mixtral} & \textbf{Yi 34B} \\
 & \textbf{13B (\%)} & \textbf{V1 (\%)} & \textbf{(\%)} & \textbf{(\%)} & \textbf{(\%)} & \textbf{(\%)} & \textbf{34B (\%)} & \textbf{(\%)} & \textbf{7B (\%)} & \textbf{8x7B (\%)} & \textbf{(\%)} \\
\midrule
Abstract2Title & 0.39 & 0.00 & 0.00 & 0.00 & 0.00 & 0.00 & 0.00 & 0.00 & 99.61 & 0.00 & 0.00 \\
ARC-Challenge  & 31.73 & 1.30 & 0.21 & 0.14 & 0.27 & 0.96 & 0.76 & 0.21 & 39.15 & 19.92 & 5.36 \\
BiasDetection  & 28.41 & 12.50 & 6.25 & 0.00 & 6.82 & 2.84 & 1.14 & 6.82 & 9.66 & 22.73 & 2.84 \\
Consensus      & 13.26 & 18.78 & 12.43 & 2.49 & 3.04 & 0.28 & 2.21 & 3.87 & 24.03 & 7.46 & 12.15 \\
GSM8K          & 25.48 & 4.20 & 1.95 & 1.34 & 4.09 & 1.44 & 5.22 & 5.09 & 28.67 & 15.15 & 7.37 \\
HellaSwag      & 13.03 & 6.14 & 0.35 & 0.71 & 3.19 & 1.46 & 0.00 & 1.74 & 26.11 & 13.29 & 33.97 \\
MBPP           & 15.41 & 3.78 & 0.54 & 1.89 & 7.84 & 2.43 & 13.24 & 0.27 & 38.38 & 14.32 & 1.89 \\
MMLU           & 30.83 & 3.28 & 1.00 & 1.01 & 2.19 & 2.95 & 0.00 & 0.00 & 26.63 & 23.02 & 9.08 \\
MT-Bench       & 11.25 & 2.50 & 0.00 & 1.25 & 12.50 & 12.50 & 7.50 & 2.50 & 20.00 & 21.25 & 8.75 \\
WinoGrande     & 27.86 & 5.84 & 0.00 & 0.00 & 1.74 & 0.71 & 1.34 & 3.25 & 21.49 & 22.61 & 15.15 \\
\bottomrule
\end{tabular}%
}
\end{table}

\subsection{Accuracy Analysis by Task}
Table~\ref{tab:routing_accuracy} reports model accuracy across benchmarks. GPT-4 emerges as a consistent all-round performer, maintaining strong results across most tasks, whereas the remaining models exhibit more variable performance across tasks depending on the benchmark.

\begin{table}[ht]
\centering
\caption{Accuracy of each model across benchmarks. Values denote the mean performance value of model over all queries across the benchmark category.}
\label{tab:routing_accuracy}
\resizebox{\textwidth}{!}{%
\begin{tabular}{lccccccccccc}
\toprule
\textbf{Benchmark} & \textbf{Mistral} & \textbf{WizardLM} & \textbf{Yi} & \textbf{Mixtral} & \textbf{CodeLlama} & \textbf{Claude Inst.} & \textbf{GPT-4} & \textbf{Claude V1} & \textbf{GPT-3.5} & \textbf{Claude V2} & \textbf{Llama} \\
 & \textbf{7B} & \textbf{13B} & \textbf{34B} & \textbf{8x7B} & \textbf{34B} & \textbf{V1} & & & & & \textbf{70B} \\
\midrule
Abstract2Title & 0.996 & 0.992 & 1.000 & 0.996 & 0.996 & 0.996 & 1.000 & 0.996 & 1.000 & 1.000 & 0.988 \\
ARC-Challenge  & 0.391 & 0.616 & 0.870 & 0.840 & 0.377 & 0.810 & 0.971 & 0.877 & 0.839 & 0.877 & 0.741 \\
BiasDetection  & 0.097 & 0.341 & 0.114 & 0.438 & 0.114 & 0.432 & 0.608 & 0.534 & 0.540 & 0.489 & 0.369 \\
Consensus      & 0.642 & 0.588 & 0.637 & 0.693 & 0.661 & 0.682 & 0.570 & 0.610 & 0.618 & 0.675 & 0.672 \\
GSM8K          & 0.412 & 0.506 & 0.548 & 0.519 & 0.457 & 0.627 & 0.659 & 0.651 & 0.605 & 0.663 & 0.523 \\
HellaSwag      & 0.261 & 0.342 & 0.761 & 0.427 & 0.213 & 0.600 & 0.860 & 0.583 & 0.601 & 0.640 & 0.539 \\
MBPP           & 0.397 & 0.427 & 0.446 & 0.624 & 0.597 & 0.697 & 0.792 & 0.689 & 0.754 & 0.741 & 0.381 \\
MMLU           & 0.266 & 0.468 & 0.690 & 0.665 & 0.005 & 0.625 & 0.850 & 0.688 & 0.678 & 0.658 & 0.028 \\
MT-Bench       & 0.524 & 0.545 & 0.644 & 0.648 & 0.528 & 0.629 & 0.806 & 0.649 & 0.689 & 0.675 & 0.586 \\
WinoGrande     & 0.524 & 0.507 & 0.629 & 0.552 & 0.384 & 0.620 & 0.819 & 0.660 & 0.579 & 0.661 & 0.482 \\

\bottomrule
\end{tabular}%
}
\end{table}

\subsection{Cost Analysis by Task}
Table~\ref{tab:cost_analysis} summarizes the total inference cost incurred by each model across benchmark tasks. As expected, open-source models are substantially more cost-efficient than their closed-source counterparts.

\begin{table*}[ht]
\centering
\small
\caption{Total cost for each model across benchmarks.}
\label{tab:cost_analysis}
 \resizebox{\textwidth}{!}{%
 \begin{tabular}{lccccccccccc}
\toprule
\textbf{Task} & \textbf{Mistral-7B} & \textbf{WizardLM-13B} & \textbf{Yi-34B} & \textbf{Mixtral-8x7B} & \textbf{CodeLlama-34B} & \textbf{Claude-Instant} & \textbf{GPT-4} & \textbf{Claude-v1} & \textbf{GPT-3.5} & \textbf{Claude-v2} & \textbf{Llama-70B} \\
\midrule
Abstract2Title & 0.015 & 0.023 & 0.061 & 0.047 & 0.058 & 0.076 & 0.923 & 0.758 & 0.081 & 0.827 & 0.104 \\
ARC-Challenge  & 0.026 & 0.039 & 0.103 & 0.078 & 0.101 & 0.106 & 1.337 & 1.057 & 0.132 & 1.057 & 0.118 \\
BiasDetection  & 0.011 & 0.019 & 0.044 & 0.035 & 0.043 & 0.056 & 1.445 & 0.636 & 0.058 & 0.894 & 0.055 \\
Consensus      & 0.021 & 0.042 & 0.073 & 0.063 & 0.091 & 0.118 & 1.219 & 0.692 & 0.111 & 1.480 & 0.105 \\
GSM8K          & 0.693 & 1.181 & 2.878 & 1.971 & 2.503 & 4.343 & 63.678 & 37.025 & 3.922 & 45.010 & 2.908 \\
HellaSwag      & 0.420 & 0.629 & 1.671 & 1.259 & 1.629 & 1.694 & 21.214 & 16.870 & 2.111 & 16.879 & 1.884 \\
MBPP           & 0.018 & 0.030 & 0.083 & 0.063 & 0.057 & 0.176 & 3.424 & 1.549 & 0.122 & 2.145 & 0.108 \\
MMLU           & 0.379 & 0.568 & 1.509 & 1.136 & 1.469 & 1.525 & 19.153 & 15.252 & 1.895 & 15.253 & 1.704 \\
MT-Bench       & 0.008 & 0.017 & 0.050 & 0.031 & 0.038 & 0.085 & 1.839 & 0.951 & 0.065 & 0.837 & 0.055 \\
WinoGrande     & 0.013 & 0.019 & 0.051 & 0.039 & 0.050 & 0.053 & 0.661 & 0.521 & 0.065 & 0.521 & 0.058 \\
\bottomrule
\end{tabular}%
}
\end{table*}

\section{Experimental Setup}
\label{appendix:experimental_setup}

\subsection{Data Splits and Preprocessing}
We partition the 34,623 filtered prompts into train (70\%), validation (15\%), and test (15\%) sets, stratified by benchmark category to ensure representative distributions. For MMLU, we further ensure balanced representation across subject areas. For the MT-Bench category, the number of available samples was relatively small which limited the model’s ability to learn a rich set of distinguishing features. To mitigate this, we kept the feature metric compact and trained it jointly with related categories, allowing shared representations to improve generalization while avoiding overfitting.

\subsection{Training Configuration}
We train LLMRank with the following hyperparameters:
\begin{itemize}[itemsep=1pt,topsep=2pt]
    \item \textbf{Architecture}: 2 hidden layers with 512 and 256 units
    \item \textbf{Optimization}: AdamW with learning rate 3e-4, weight decay 1e-2
    \item \textbf{Training}: 100 epochs, batch size 256, early stopping with patience 10
    \item \textbf{Regularization}: Dropout 0.2, gradient clipping at norm 1.0
\end{itemize}

\subsection{Training Configuration}

Table~\ref{tab:training_config} summarizes the hyperparameters used for training LLMRank.  

\begin{table}[ht]
\centering
\caption{Training configuration for LLMRank.}
\label{tab:training_config}
\begin{tabular}{ll}
\toprule
\textbf{Component} & \textbf{Setting} \\
\midrule
Architecture & 2 hidden layers (512, 256 units) \\
Optimizer & AdamW (lr = 3e-4, weight decay = 1e-2) \\
Training & 100 epochs, batch size 256 \\
Early stopping & Patience 10 \\
Regularization & Dropout 0.1, gradient clipping at norm 1.0 \\
Sentence transformer & all-MiniLM-L6-v2 \\
\bottomrule
\end{tabular}
\end{table}

We train three variants of LLMRank by varying the utility trade-off parameter $\lambda$:  
\begin{itemize}[itemsep=1pt,topsep=2pt]
    \item \textbf{Performance-first}: $\lambda = 0$ (prioritize quality)
    \item \textbf{Balanced}: $\lambda = 10^3$ (balance quality and cost)
    \item \textbf{Cost-first}: $\lambda = 10^5$ (prioritize cost savings)
\end{itemize}

\subsection{Baselines}
We compare against several routing strategies:
\begin{itemize}[itemsep=1pt,topsep=2pt]
    \item \textbf{Oracle}: Perfect router that selects optimal model per prompt
    \item \textbf{Best Single}: Always use the single best model (GPT-4)
    \item \textbf{Cheapest}: Always select the lowest-cost model (Mistral-7B)
\end{itemize}

\subsection{Evaluation Metrics}
\begin{itemize}[itemsep=1pt,topsep=2pt]
    \item \textbf{Efficiency}: Ratio of router and oracle quality: $\text{Efficiency} = Q_{\text{router}}/Q_{\text{oracle}}$
    \item \textbf{Cost Ratio}: Ratio of router and oracle cost: $\text{Cost Ratio} = C_{\text{router}}/C_{\text{oracle}}$
    \item \textbf{Quality Gap}: Difference of oracle and router quality: $\text{Efficiency} = Q_{\text{oracle}} - Q_{\text{router}}$
\end{itemize}

\section{Example Routing Decisions}

We provide one concrete example from our routing analysis to illustrate LLMRank's behavior on actual RouterBench prompts.:

\subsection{Example 1: Arc-Challenge (sample id: arc-challenge.test.1011)}
\begin{verbatim}
Prompt: "Which question would scientists studying prokaryotic 
organisms most likely ask?
A) How do lysosomes expel bacteria from cells?
B) What role does the cell membrane play in stability?
C) Why are ribosomes efficient protein producers?
D) How do chloroplasts convert light into energy?
Print only a single choice from 'A' or 'B' or 'C' or 'D' without explanation.
Answer:"
Features: {
task_type: "biology_multiple_choice",
complexity_score: 0.40,
reasoning_steps: 2,
domain: "cell_biology"
...
}
Oracle Model: mixtral-8x7b-chat
Selected Model: mixtral-8x7b-chat
Reasoning: Requires domain-specific biology knowledge to identify features 
of prokaryotic organisms and exclude distractors related to eukaryotic 
organelles.
Quality: 1.0 (correct: C)
\end{verbatim}

\end{document}